\pdfoutput=1

\documentclass[11pt]{article}

\usepackage[preprint]{coling}

\usepackage{times}
\usepackage{latexsym}

\usepackage[T1]{fontenc}

\usepackage[utf8]{inputenc}

\usepackage{microtype}

\usepackage{inconsolata}

\usepackage{graphicx}
\usepackage{multirow}
\usepackage{enumitem}
\usepackage{adjustbox}
\usepackage[para]{footmisc}
\usepackage{tabularx}
\usepackage{hyperref}
\usepackage{url}
\usepackage{booktabs}
\usepackage{amsfonts}
\usepackage{nicefrac}
\usepackage{xcolor}

\usepackage{amsmath}
\usepackage{amssymb}
\usepackage{amsthm}
\usepackage{subcaption}
\usepackage{listings}
\usepackage{siunitx}

\usepackage{pifont}
\usepackage[flushleft]{threeparttable}

\title{LLM-Based Insight Extraction for Contact Center Analytics and Cost-Efficient Deployment}

\author{
  Varsha Embar \qquad Ritvik Shrivastava \qquad Vinay Damodaran \\  \textbf{Travis Mehlinger \qquad
  Yu-Chung Hsiao \qquad Karthik Raghunathan}
\\
\\
 Cisco Systems
 \\ \\
 \normalsize{\texttt{\{vembar, ritvshri, vidamoda, tmehling, yohsiao, ktick\}@cisco.com}}
}

\begin{document}
\maketitle
\begin{abstract}

Large Language Models have transformed the Contact Center industry, manifesting in enhanced self-service tools, streamlined administrative processes, and augmented agent productivity.
This paper delineates our system that automates call driver generation, which serves as the foundation for tasks such as topic modeling, incoming call classification,   trend detection, and FAQ generation, delivering actionable insights for contact center agents and administrators to consume.
We present a cost-efficient LLM system design, with 1) a comprehensive evaluation of proprietary, open-weight, and fine-tuned models and 2) cost-efficient strategies, and 3) the corresponding cost analysis when deployed in production environments.

\end{abstract}

\section{Introduction}

The Contact or Call Center (CC) industry has rapidly adopted Large Language Models (LLMs) to extract insights from call streams to efficiently resolve customers' problems. 
The recent improvements in LLM quality, performance, and accessibility have facilitated their integration into Interactive Voice Response (IVR) systems and virtual agents or chatbots. 
This integration has two main objectives: 1) to boost containment rates\footnote{Containment rate refers to the fraction of calls handled successfully without human intervention}, and 2) to shorten agent handle times by offering real-time assistance via suggested responses linked to relevant documents during calls.
In both cases, recognizing topics that customers frequently call about is a critical prerequisite to optimize CC operations.


Traditional topic modeling systems provide valuable insights but are limited by their dependence on transcript lexicons, often producing overlapping topics and multiple keywords from a single call. This ambiguity complicates interpretation and typically requires manual curation by CC administrators to meet business needs.
To address these limitations, our approach shifts from comprehensive call topics to extracting a customer's primary contact reason, referred to as a \textit{call driver}, aiming to reduce ambiguity and improve interpretability in downstream analyses.
We leverage the generalization capabilities of instruction-tuned LLMs as our modeling foundation to alleviate 
the challenges in data availability and human annotation at scale. 
While some vendors' generic LLMs can generate quality call drivers via prompt engineering, exclusive reliance on third-party solutions presents several drawbacks: 1) escalated costs due to high call volumes, 2) privacy concerns associated with sending sensitive data externally, and 3) prohibitive expenses for provider managed fine-tuning when applications require extra customization.

In this paper, we 
present an LLM system for a commercial offering that provides contact center insights with the goal to improve the productivity metrics. 
The key modules are as follows:

\begin{itemize}[leftmargin=*]
    \itemsep-4px
    \item Call driver generation (\S\ref{call_driver}): Extract concise call drivers as primary representations of transcripts, serving as input for all downstream applications. 
    \item Topic modeling (\S\ref{sec:topic_modeling}): Utilize call drivers to derive topic clusters with appropriate labels.
    \item Trend detection (\S\ref{sec:trend_detection}): Identify new and emerging topics by maintaining sub-topics derived from our topic models.
    \item Frequently Asked Questions (FAQs) generation (\S\ref{sec:faqs}): Use call drivers from frequent topics to identify FAQs. 
\end{itemize}

Finally, we demonstrate how efficient fine-tuning, input compression, and deployment strategies can reduce LLM production costs compared to employing proprietary vendors, offering a cost-effective solution for organizations seeking to improve their contact center operations.


\section{Related Work}
The advent of transformer-based architectures \citep{vaswani2017attention} led to significant improvements in the Contact Center~(CC) industry. 
Popular CC applications include analyzing mentioned entities with customer sentiment~\citep{fu2022entity, tahmid2022improving}, 
generating a post call summary~\citep{zou2021topic}, and building virtual agents that provide responses grounded in retrieved documents~(RAG)~\citep{bonetta2021retrieval}. 
Instruction-tuned models have expanded these applications. Commercial offerings provide LLM-powered call analytics, features to assist agents during calls and enable data-driven decisions.\footnote{\href{https://cloud.google.com/solutions/ccai-insights}{Google Cloud CCAI Insights}}\footnote{\href{https://aws.amazon.com/machine-learning/ml-use-cases/contact-center-intelligence/}{AWS CC Intelligence}}. 

Understanding the customers' call reasons allows CC admins to equip agents with the necessary training resources to satisfy customers' needs~\citep{khasanova2022developing}. 
Automatic call classification with predetermined classes has also been explored~\citep{tang2003call}. 
When predetermined classes are unavailable,
traditional topic modeling approaches are employed \citep{papadia2022topic, hendry2021topic}. 
However, a single call may be assigned with multiple labels, due to certain keyphrases.
This ambiguity hinders downstream analyses from capturing the essence of the call. 
\section{Call Driver Generation}
\label{call_driver}
\begin{figure}[ht]
    \centering
    \includegraphics[width=\columnwidth]{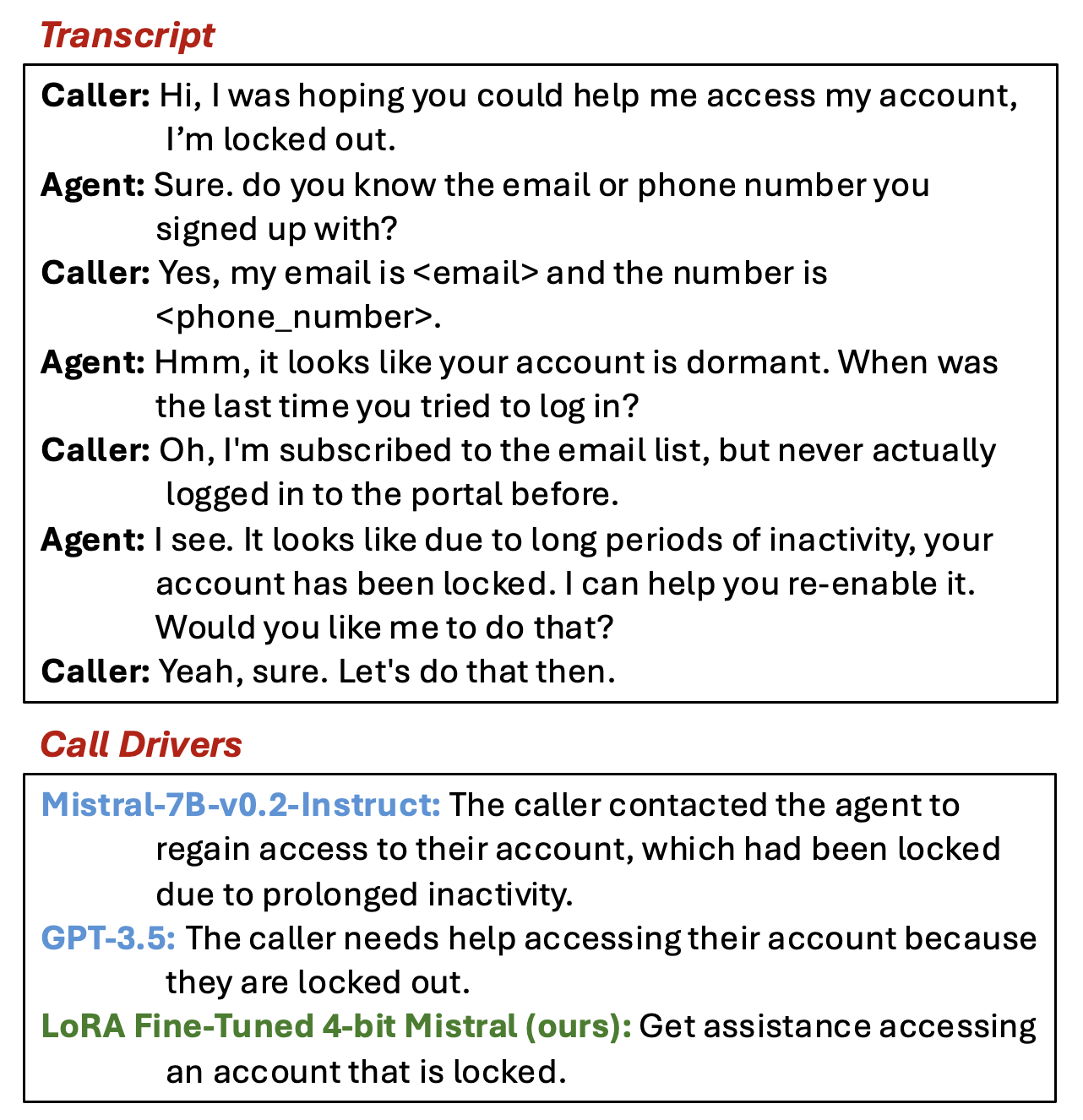}
    \caption{A sample call transcript with call driver generations from various models.}
    \label{fig:types_of_drivers}
\end{figure}

A \textit{call driver} is a concise 15-20 word description summarizing the primary reason for a customer's call.
Figure \ref{fig:types_of_drivers} provides an illustrative example.
Our objective is to develop a production-ready model that extracts these main reasons from call transcripts, considering various factors including privacy, legal compliance, cost efficiency, and latency optimization. 
The resulting call drivers will be further consumed by downstream tasks detailed in~\S\ref{sec:cc_insights}.

\subsection{Datasets \& Evaluation} \label{sec:call_driver_datasets_evaluation}
\paragraph{Test Set}
To evaluate the model-generated call drivers, we use two internal real-world CC datasets from shipping (2k calls) and IT help-desk (3k calls) domains. We transcribe these audio datasets using Azure\footnote{\href{https://azure.microsoft.com/en-us/products/ai-services/speech-to-text}{Azure Speech-to-Text}} to get 
diarized transcripts. 
These transcripts are then human annotated by one professional annotator per call to establish ground truth call drivers.

\paragraph{Training Set}
To comply with our CC product's privacy policy, we generate 750 synthetic transcripts through human role-playing, avoiding use of customer data for model training. Two annotators simulate calls - as caller and agent - based on a provided domain, problem, and potential solution. A separate set of annotators label the call drivers.

\paragraph{Evaluation}
The shortcomings of standard metrics for evaluating model generated text with ground-truth references like ROUGE, BertScore etc., is well-researched \cite{fabbri2021summeval, liu2022revisiting}. 
Previous work \cite{khobragade2019machine} has shown that entailment of text correlates better with human judgement in detecting paraphrases. 
However, entailment models trained on Natural Language Inference (NLI) datasets demonstrate a preference for longer hypotheses \citep{gururangan2018annotation, guo2023length}. 
Inspired by \citep{papineni2002bleu,dubois2024length}, we introduce a length penalty as a remedy.

We use
\texttt{nli-deberta-v3-base}\footnote{\href{https://www.sbert.net/docs/pretrained_cross-encoders.html\#nli}{SBERT-NLI}} as our entailment model denoted as $\textrm{entails}(\cdot)$. 
For the $i$-th pair of reference ($ref_i$) and hypothesis ($hyp_i$) of a dataset of size $n$, we classify the pair as positive (entailment or neutral) or negative (contradiction): 
\begin{align} \label{eq:call_driver_metric}
\begin{split}
    l_p &= \min\left(1, \ \alpha\sqrt{\frac{\sum_i \mbox{len}(ref_i)}{\sum_i \mbox{len}(hyp_i)}} \right) \\
    S_{\textrm{cd}} &= l_p \times \tfrac{1}{n} \sum_i \mbox{entails}(ref_i, hyp_i),
\end{split}
\end{align}
where $l_p$ is the corpus length penalty, $\alpha$ is a fixed scaling factor, and $\textrm{len}(\cdot)$ computes the length of text.
The corpus length penalty $l_p$
is introduced for two reasons: 1) to encourage the generation to focus on a single, or fewer most important call reasons, to disambiguate the downstream tasks for CC insights in \S\ref{sec:cc_insights}, and 2) to empirically calibrate the call driver score $S_{\textrm{cd}}$ to the end-to-end score $S_{\textrm{e2e}}$ in~\eqref{eq:topic_model_metric}.
This allows the iteration of call driver modeling to be loosely decoupled from end-to-end topic modeling evaluation, to accelerate the experimental time.
The scaling factor $\alpha$ is determined to be~1 during our end-to-end topic modeling experiments.

\subsection{Zero-shot baselines and fine-tuned models}
\paragraph{Zero-shot baseline} 
We selected two language models as baselines for call driver generation: the proprietary \texttt{gpt-3.5-turbo} and the open-weights \texttt{Mistral-7B-Instruct-v0.2}. 
Each model utilizes a custom prompt design. 



\paragraph{Fine-tuned model} In addition, we fine-tune \texttt{Mistral-7B-Instruct-v0.2} with LoRA \citep{hu2021lora} on our synthetic dataset, employing a 70:30, train:validation split. To reduce the model size, we use 4-bit quantization aware LoRA training \citep{frantar2022gptq} with rank 64. We opt for LoRA as opposed to full model fine-tuning to enable both call driver and topic label generation (in \S\ref{sec:topic_modeling}) using the same backbone model as part of our efficient strategy, as further described in~\S\ref{sec:cost_efficient_strategies}.

\subsection{Results}

\begin{figure}[t]
    \centering
    \includegraphics[width=0.8\columnwidth]{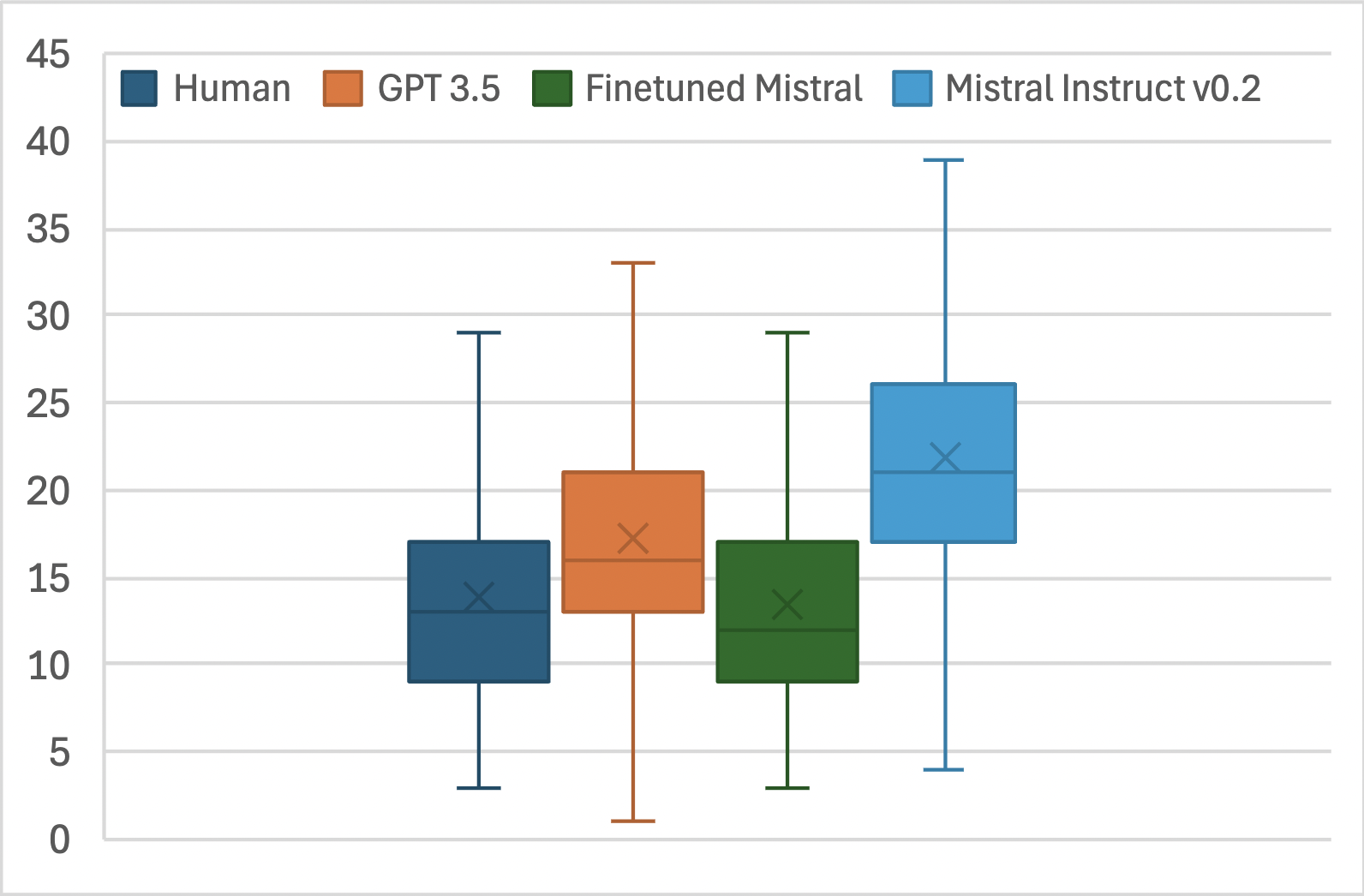}
    \caption{Call driver length distributions reveal notable differences among models. The zero-shot baselines (GPT and Mistral) tend to generate longer call drivers, while our fine-tuned model align closely with human annotations, despite being trained on a separate synthetic dataset. Further analysis indicates that longer call drivers often include multiple detailed call reasons and are more likely to be rated as entailment neutral. This negatively impacts end-to-end performance (Table~\ref{tab:results}). }
    \label{fig:len_distrib}
\end{figure}

We evaluated the three models against the human annotated test set. The results are summarized in Table \ref{tab:results}.
The corresponding call driver word length distributions are depicted in Figure~\ref{fig:len_distrib}.
We observe that the baseline models tend to generate longer outputs containing multiple call reasons in a single call driver. 
For instance, in ascending order:

{\small
\begin{itemize}[leftmargin=*]
    \itemsep-2px
    \item Our fine-tuned model: ``To request a loaner laptop.''
    \item GPT-3.5 baseline: ``The caller requested an additional laptop and wanted to know the procedure for requesting it.''
    \item Mistral baseline: ``Caller requested information on how to request an additional laptop with specifications and inquired about the approval and delivery process.''
\end{itemize}
}

\noindent These longer outputs not only bias the entailment models to classify them as neutral, but also negatively impact end-to-end performance in tasks described in \S\ref{sec:cc_insights}. 
Therefore, developing a model capable of generating concise, targeted call drivers is crucial for enhancing performance for these tasks.

\begin{table*}[t]
\centering
\footnotesize{
\begin{tabular}{lcccccccc}
\toprule
\multicolumn{1}{c}{\multirow[b]{2}{*}{\raisebox{2pt}{Model}}} & \multicolumn{2}{c}{Call Driver w/o $l_p$ \eqref{eq:call_driver_metric} $\uparrow$} & \multicolumn{2}{c}{Call Driver (\S\ref{call_driver}) $\uparrow$} & \multicolumn{2}{c}{E2E (\S\ref{sec:topic_modeling}) $\uparrow$} & \multicolumn{2}{c}{DBCV (\S\ref{sec:topic_modeling}) $\downarrow$}\\ 
\cmidrule(lr){2-3}
\cmidrule(lr){4-5}
\cmidrule(lr){6-7}
\cmidrule(lr){8-9}
\multicolumn{1}{c}{} & \multicolumn{1}{c}{Shipping} & \multicolumn{1}{c}{IT} & \multicolumn{1}{c}{Shipping} & \multicolumn{1}{c}{IT} & \multicolumn{1}{c}{Shipping} & \multicolumn{1}{c}{IT} & \multicolumn{1}{c}{Shipping} & \multicolumn{1}{c}{IT} \\ 
\midrule
Test Set Human Annotations & -- & -- & -- & -- & -- & -- & 0.54 & \underline{0.46} \\
\cmidrule(lr){1-1}
GPT 3.5  & \textbf{94.60} & \textbf{95.97} & \underline{84.03}  & \textbf{85.91}  & \underline{81.91}  & \underline{80.54} & \underline{0.52} & 0.51  \\
Mistral 7B Instruct v0.2  & 89.27 & \underline{92.97} & 74.44 & 73.46  & 77.79  & 80.43 & 0.63 & 0.55 \\
\midrule
LoRA FT 4-bit Mistral (ours) & \underline{89.70} & 86.06 & \textbf{88.88} & \underline{85.04}  & \textbf{82.66}   & \textbf{83.00} & \textbf{0.23} & \textbf{0.44} \\ 
\bottomrule
\end{tabular}
}
{%
    \caption{
        Evaluation of Shipping and IT Helpdesk across all metrics. The best scores are in bold while the 2nd are underlined. 
        The Mistral baseline tends to generate significant longer call drivers (see Figure~\ref{fig:len_distrib}) that degrade the DBCV and E2E performance.
        The corpus length penalty $l_p$~\eqref{eq:call_driver_metric} calibrates the Call Driver scores to facilitate model selections without the need of evaluating the end-to-end topic modeling pipeline. 
        Note that the clustering quality is better when the DBCV value is lower. Other metrics are the higher the better.
    }
    \label{tab:results}
}
\end{table*}
 
\section{Automated Contact Center Insights}
\label{sec:cc_insights}

\subsection{Topic Modeling}
\label{sec:topic_modeling}

\begin{figure}[h]
    \centering
    \includegraphics[width=0.9\columnwidth]{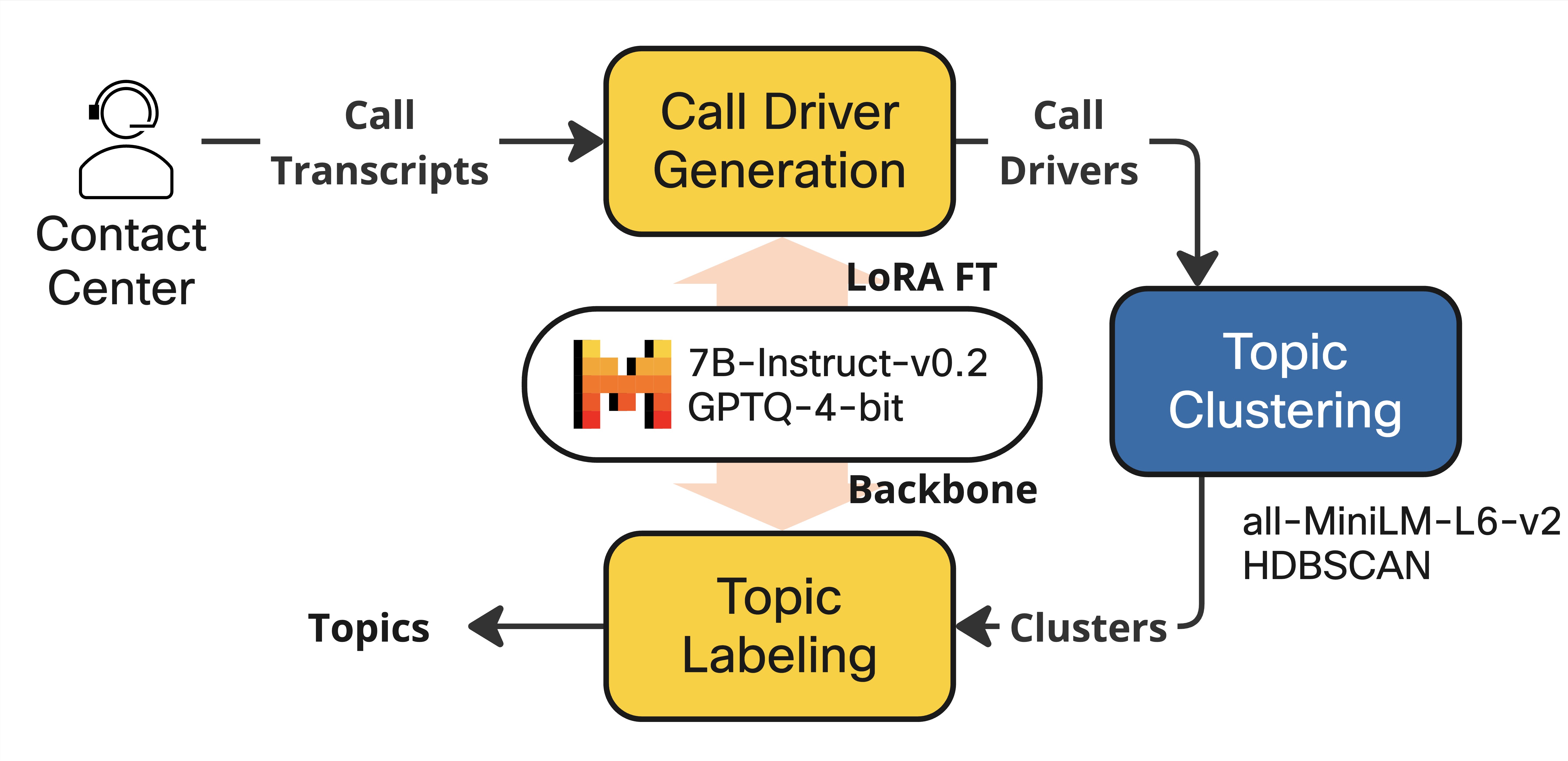}
    \caption{Topic modeling pipeline. A single Mistral model is deployed for both Call Driver Generation (LoRa fine-tuned) and Topic Labeling (backbone) as part of our cost-efficient strategy.}
    \label{fig:topic_modeling_pipeline}
\end{figure}

Contact centers (CCs) handle a large number of calls daily. Identifying the most common and recurring topics helps CC administrators in resource allocation and prioritization. Using generated call drivers as our seed set, we employ a multi-step topic modeling algorithm. 
The overall pipeline is illustrated in Figure~\ref{fig:topic_modeling_pipeline}.
We embed call drivers using \texttt{all-MiniLM-L6-v2} \cite{reimers-2019-sentence-bert}. We use HDBSCAN \citep{mcinnes2017hdbscan} to generate topic clusters from call driver embeddings. We perform a grid search to determine the optimal hyperparameters for each data set, using the DBCV score \citep{moulavi2014density} as the optimization metric. 
HDBScan was selected and preferred over alternative clustering algorithms such as K-means and Gaussian Mixture Models because it does not require a predefined number of clusters/topics. 

After clustering, the final step is to label the clusters, serving as \textit{topics} for the end user. 
We use the same 4-bit Mistral LLM for Call Driver Generation in~\S\ref{call_driver} but without the fine-tuned adapter (referred to as the \textit{backbone} model). 
For each cluster, the following few-shot prompt is utilized to generate a topic label, using two inputs: 1) a subset of call drivers from the cluster, and 2) the $k$-most common keywords associated with the cluster:

{\small
\begin{verbatim}
[INST]  Generate a title in up to five words
for the following phrases: {}; 
and most common words: {}.  [/INST]
\end{verbatim}
}

\noindent 
To select representative cluster members and keywords, we preprocess call drivers with stop-word removal and lemmatization. 
For each cluster, we identify the top-$n$ most frequent normalized drivers and select one original call driver corresponding to each as the cluster representatives.
Common keywords for a cluster are derived from the top-$k$ most frequent unigrams after normalization. Optimal values of $n=25$ and $k=3$ were determined through a hyperparameter search.
Table \ref{tab:topic_modeling} shows some representative topics with their labels and corresponding call drivers from the synthetic dataset. 


\begin{table*}[ht]
\centering
\begin{adjustbox}{width=\textwidth}
\setlength{\tabcolsep}{2pt}
\begin{tabular}{lc}
\toprule
\multicolumn{1}{c}{\textbf{Sample Call Drivers}} & \multicolumn{1}{c}{\textbf{Topic Label}} \\ 
    \midrule
    \renewcommand{\arraystretch}{0.8}
    \begin{tabular}{p{1.05\textwidth}}
        \setlist{nolistsep}
        \begin{enumerate}
            \item[1.1] Get assistance assessing current retirement planning and savings and discussing ways to improve.
            \item[1.2] Get assistance creating a retirement plan that ensures financial stability for a longer period.
            \item[1.3] Get assistance figuring out a plan to avoid outliving retirement savings.
        \end{enumerate}
        \end{tabular}                   
    & Discussing Retirement Planning Challenges \\ 
    \cmidrule(r){1-1} \cmidrule{2-2}
    \renewcommand{\arraystretch}{0.8}
    \begin{tabular}{p{1.13\textwidth}}
    \setlist{nolistsep}
    \begin{enumerate}
        \item[2.1] Discuss an issue with inaccurate flight information and receive assistance booking on the next available flight.
        \item[2.2] Asking for assistance in changing a flight.
        \item[2.3] Asking for assistance with a wicked problem with flight booking and ticketing for a family vacation.
    \end{enumerate} 
    \end{tabular}          
    & Resolving Flight Update Issues \\
    \bottomrule
\end{tabular}
\end{adjustbox}
{%
    \caption{
        Example call drivers and topic labels generated by our LoRA FT 4-bit Mistral on the synthetic data.
    }%
    \label{tab:topic_modeling}
}
\vspace{-5mm}
\end{table*}

\paragraph{Evaluation} 
We use DBCV scores to assess cluster cohesion (Table~\ref{tab:results}). In addition, we also evaluate coherence between the generated topic labels and their associated call drivers: 
For each cluster, we compute two scores: 1) \textbf{Cosine similarity}: average cosine similarity between embeddings of the label and associated call drivers, and 2) \textbf{Entailment}: average entailment scores from the label to the associated member call drivers.
The resulting scores are averaged across clusters.
We use the models \texttt{all-MiniLM-L6-v2} for embedding and \texttt{nli-deberta-v3-base} for entailment computation.
Formally, for a set of $n$ clusters, in which the $i$-th cluster, labeled as $lb_i$, contains $m_i$ associated call drivers $cd_j$, the end-to-end (E2E) score $S_\textrm{e2e}$ is computed as:
\begin{align} \label{eq:topic_model_metric}
\begin{split}
    S_{\textrm{sim}} &= \tfrac{1}{n}\sum_{i=1}^{n} \tfrac{1}{m_i} \sum_{j=1}^{m_i}\cos(lb_{i}, cd_{j}) \\
    S_{\textrm{ent}} &= \tfrac{1}{n}\sum_{i=1}^{n} \tfrac{1}{m_i} \sum_{j=1}^{m_i} \textrm{entails}(lb_{i}, cd_{j}) \\
    S_{\textrm{e2e}} &= \tfrac{1}{\alpha + \beta} \ (\alpha  S_{\textrm{sim}} + \beta  S_{\textrm{ent}}),
\end{split}
\end{align}
where $\alpha$ and $\beta$ are given weights.
Note that the metric design of incorporating both similarity and entailment is to balance between semantic relevance and logic coherence within a topic.
For e.g., a cluster label ``Overdraft Insurance Enrolment'' is similar to a call driver ``Wants to opt out of overdraft insurance'' but rejects that by entailment. 
The choice of the corresponding weights $\alpha$ and $\beta$ therefore is business driven and can vary according to the industry domains.
We found that a simple average is sufficient for the two domains in our study.


\paragraph{Results}
Table \ref{tab:results} shows evaluation scores for the 3 models. 
We observed that long call drivers (Figure~\ref{fig:len_distrib}) frequently contain excess details, resulting in formation of generic, non-informative clusters, with labels that often fail to accurately represent all call drivers within a cluster.
This supports the need of introducing length penalty in~\eqref{eq:call_driver_metric}.

\subsection{Call Classification}
A critical application of topic modeling lies in dynamic categorization of incoming calls without necessitating model training. 
Each new call is assigned a topic label based on its proximity to the closest existing cluster within a topic model. 
Each new call that matches the closest existing cluster within the topic model is assigned with the topic label.
For example, a call driver ``Customer inquired about ordering equipment'' would be assigned to the ``Equipment Ordering'' cluster based on semantic similarity.
We monitor the frequency of new call drivers mapped to each topic. 
When a new call driver exhibits lexical divergence from existing cluster members while still fits within the context, it is merged into the topic cluster.

\subsection{New Trend Detection}
\label{sec:trend_detection}

A topic model generates clusters of different sizes — some large and prominent, followed by smaller clusters and an outlier catch-all cluster. A new call can be assigned to 1) a prominent topic cluster, 2) a smaller topic cluster, or 3) the outlier cluster.

If the second scenario occurs frequently, the smaller cluster will grow to resemble the larger clusters, signaling the emergence of a new trend. 
In contrast, a growing outlier cluster does not clearly indicate a well-defined emerging topic. 
Re-clustering of the outlier cluster members can be performed using a fast, greedy clustering algorithm\footnote{\href{https://github.com/UKPLab/sentence-transformers/blob/master/examples/applications/clustering/fast_clustering.py}{Fast-Clustering - Sentence Transformers}}. 
New call drivers are classified into these sub-clusters. If no match is found, the new call driver is added as a single-element cluster within the outlier group. This allows us to track emerging trends within the outlier cluster.

This approach keeps the model updated without re-training the entire topic model for every new batch of calls, significantly reducing the overhead.

\subsection{Frequently Asked Questions}
\label{sec:faqs}
Frequently asked questions (FAQs) often reveal recurring issues with products or services, signaling areas for improvement. By incorporating the FAQs into IVR flows, we can alleviate pressure on agents and allow them to focus on more complex queries. 

We approach this by extending topic modeling. For each topic cluster and its associated call drivers, we trace the drivers back to their originating utterances in transcripts using lexical overlap density scores. If multiple utterances share the same scores, we include them all as potential matches. From this pool, we randomly select 5-20 utterances and use GPT-3.5 to identify the common questions within them, and formatting them as FAQs. 
We observed that narrowing the focus to a smaller set of utterances produces higher quality FAQs, compared to using
an entire transcript or a broad sample of call drivers.
The resulting questions are then clustered for the entire set of historical calls, using a high similarity threshold, and the centroid of each cluster is selected as the representative question candidate.

\section{Cost-Efficient Strategies} \label{sec:cost_efficient_strategies}

We discuss our strategies to efficiently deploy the overall topic modeling pipeline to production.

\subsection{Multi-LoRA for Topic Modeling} \label{sec:multi-lora_topic_modeling}

Our topic modeling pipeline, depicted in Figure~\ref{fig:topic_modeling_pipeline}, employs two LLMs: one for Call Driver Generation and the other for Topic Labeling. 
We leverage MultiLoRA inference, enabling dynamic  switching between LoRA adapters, to efficiently deploy a single model that serves both purposes.

\subsection{Input Compression} \label{sec:input_compression}
Leveraging the framework LLMLingua2 \cite{pan-etal-2024-llmlingua}, we explore input compression as a further optimization path. Here, we reduce the size of call transcripts before processing text through LLMs, thereby lowering token counts and computational costs. Using a token classification model, each token is assigned a discard probability.
Tokens are then removed from the transcript up to a predetermined threshold, retaining the top-$n$ most relevant tokens. 
Our experiments demonstrate that transcript compression significantly reduces input size, costs, and latency while minimally impacting the quality of generated call drivers (Table \ref{tab:compression}). 

\begin{table}[t]
\centering
\small
\begin{tabular}{rrcc}
\toprule
\multirow{2}{*}[-11px]{Input\%} & \multirow{2}{*}[-6px]{\begin{tabular}[c]{@{}r@{}}Compres- \\ sion ratio\end{tabular}}
& \multicolumn{2}{c}{Call Driver Score (\S\ref{call_driver}) $\uparrow$} \\ 
\cmidrule(lr){3-3} \cmidrule(lr){4-4} 
& & \multicolumn{1}{c}{Shipping} & IT Helpdesk \\ 
\midrule
100\% & 1x & \multicolumn{1}{c}{\textbf{88.88}} & \textbf{85.04} \\ 
70\% & 1.4x & \multicolumn{1}{c}{\underline{88.85}} & 83.77 \\ 
50\% & 2x & \multicolumn{1}{c}{87.60} & \underline{84.09} \\ 
33\% & 3x & \multicolumn{1}{c}{84.10} & 83.12 \\ 
25\% & 4x & \multicolumn{1}{c}{84.10}& 81.04 \\ 
20\% & 5x & \multicolumn{1}{c}{84.30} & 81.63 \\ 
\bottomrule
\end{tabular}%
\caption{Call driver quality measured against varying compression ratios. Notably, multifold input token reduction still yield strong call driver scores, indicating that fine-tuned LLMs can produce quality outputs even with compressed, grammatically terse inputs.}
\label{tab:compression}
\end{table}

\subsection{Model Hosting Infrastructure and Cost} \label{sec:model_hosting}

Hosting LLMs on EKS infrastructure enables building a scalable, efficient, and cost-effective environment for serving models. At the core, we use vLLM \cite{kwon2023efficientmemorymanagementlarge} as a model server to maximize throughput, ensuring optimal performance and low-latency responses through efficient handling of concurrent requests. 
To optimize resource management, we leverage Karpenter\footnote{\href{https://karpenter.sh}{Karpenter}} as a Kubernetes provisioner, which allows us to run our infrastructure on spot instances. A single model deployment runs comfortably on an A10G GPU with 24GB memory.

\begin{figure}[t]
    \includegraphics[width=\columnwidth]{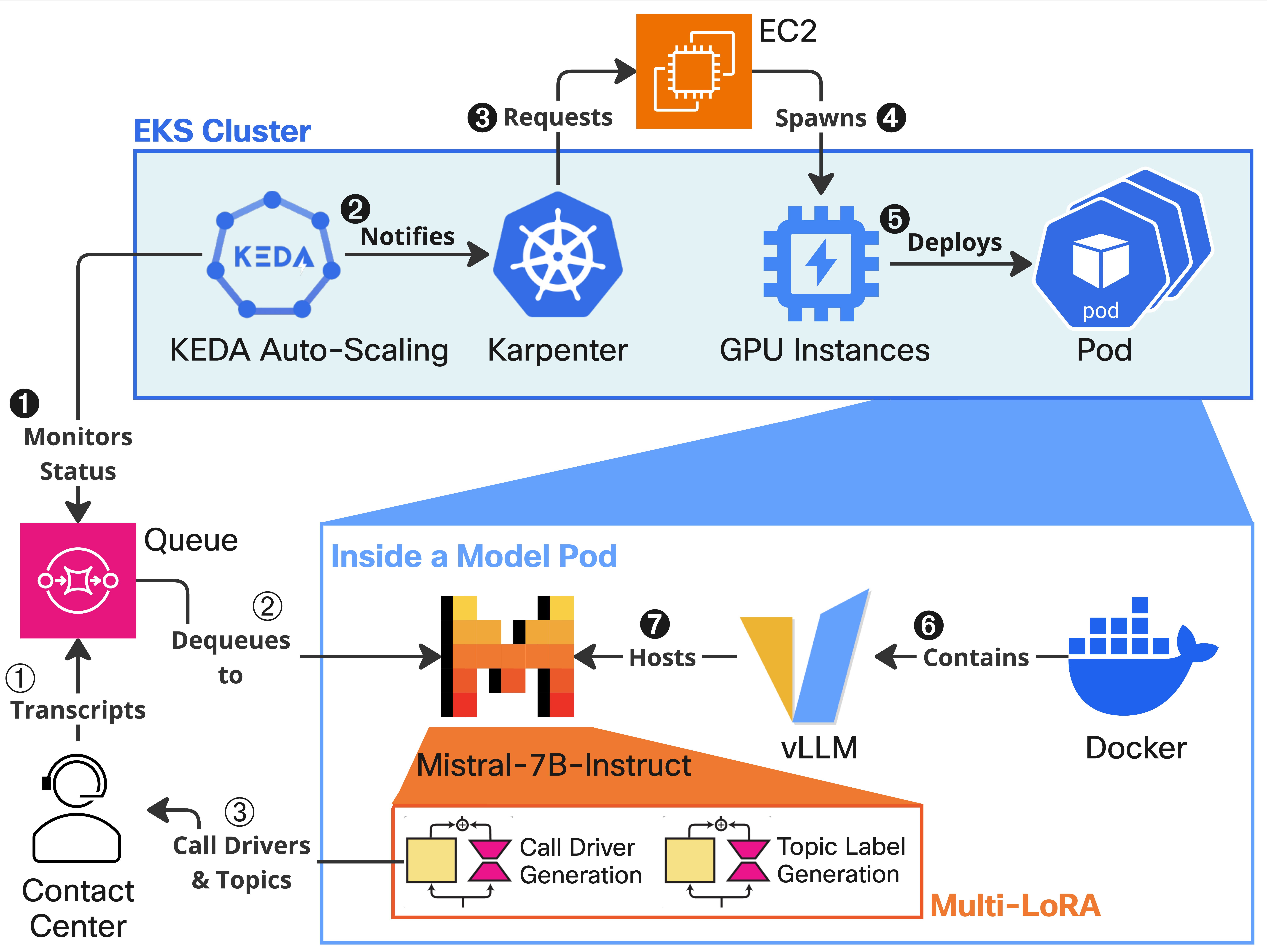}
    \caption{LLM hosting architecture. Circled numbers~\ding{192}: Steps for model inference. Filled circled numbers~\ding{203}: Steps to scale and host LLM models. KEDA monitors Queue workloads and triggers Karpenter to provision new GPU instances. This design allows us to scale up from and down to zero instances and prioritize spot over on-demand instances for the cost consideration.}
    \label{fig:infra}
\end{figure}

The architecture (Figure \ref{fig:infra}) also incorporates dynamic scalability using KEDA\footnote{\href{https://keda.sh}{KEDA}} scalers, which poll Amazon SQS
queues to trigger workloads, enabling us to scale GPU resources from zero, minimizing idle time and associated costs. Workload orchestration is handled seamlessly within Kubernetes (k8s)\footnote{\href{https://kubernetes.io}{Kubernetes}} pods, where vertical scaling based on load ensures that server resources can be dynamically adjusted according to demand. To further reduce startup times and make our Docker images lean, model weights are fetched from Amazon S3 using the s5 utility\footnote{\href{https://github.com/peak/s5cmd}{s5}}, and stored in NVMe
storage for faster access. This combination of technologies --- vLLM for throughput, LoRA for cost-efficiency, and dynamic scaling with KEDA --- creates an efficient, flexible, and cost-optimized pipeline for EKS hosting and managing of large-scale LLM models.

\begin{table}[ht]
\begin{threeparttable}
\footnotesize
\begin{tabularx}{\columnwidth}{lrr}
\toprule
Model & Cost (USD \$) & Cost Ratio \\
\midrule
\textit{\textbf{Ours}} & &\\
Lower-bound (spot) & \textbf{1.98} & \textbf{1x}\\ 
Upper-bound (on-demand) & \textbf{4.77} & \textbf{2.4x}\\
\midrule
\textit{\textbf{Proprietary Pricing}} & \\
Mistral-7B (AWS Bedrock) & 10.38 & 5.2x\\
GPT-3.5-Turbo & 14.20 & 7.2x\\
GPT-4o\tnote{$\dagger$} & 142.00 & 71.7x\\
GPT-4o-mini\tnote{$\dagger$} & \underline{4.82} & \underline{2.4x}\\
\bottomrule
\end{tabularx}
\begin{tablenotes}
    \item[$\dagger$] Models are limited to certain regions.
\end{tablenotes}
\caption{Cost comparison for \textbf{LoRA Fine-Tuned 4-bit Mistral (ours)} with proprietary vendors' token-based pricing. Benchmarking on 500k concurrent transcripts.
}
\label{tab:cost}
\end{threeparttable}
\end{table}

Table \ref{tab:cost} presents a comparative cost analysis of our EKS-hosted LLMs versus per-token pricing from proprietary vendors. We evaluated the concurrent cost for processing 500k input transcripts. The lower bound, or best-case cost, occurs when the EKS-hosted model operates exclusively on spot instances when available, while upper bound, or maximum cost, reflects an all on-demand instance scenario. Both cost bounds are significantly lower than token-based pricing models. 
The cost of GPT-4o-mini approaches our upper bound.
However, it is still regionally limited at the time of this writing and not suitable for our production purpose.

\section{Conclusion}
LLMs have accelerated application development for contact center operations. 
We presented a cost-efficient LLM system that automates insight extraction from customer call transcripts, with a primary focus on generating call drivers.
These call drivers serve as a crucial foundation for monitoring trends and topics in customer interactions.
Our case study highlights fine-tuning, hosting, and optimizing open-weight models such as Mistral Instruct v0.2, compared with proprietary models like GPT. 
We also explored how these solutions enhance downstream applications, such as topic modeling and FAQ generation, with the potential to improve call handling times, reduce human effort, and increase contact center efficiency.


\bibliography{custom}

\begin{thebibliography}{23}
\providecommand{\natexlab}[1]{#1}

\bibitem[{Bonetta et~al.(2021)Bonetta, Cancelliere, Liu, and Vozila}]{bonetta2021retrieval}
Giovanni Bonetta, Rossella Cancelliere, Ding Liu, and Paul Vozila. 2021.
\newblock Retrieval-augmented transformer-xl for close-domain dialog generation.
\newblock \emph{arXiv preprint arXiv:2105.09235}.

\bibitem[{Dubois et~al.(2024)Dubois, Galambosi, Liang, and Hashimoto}]{dubois2024length}
Yann Dubois, Bal{\'a}zs Galambosi, Percy Liang, and Tatsunori~B Hashimoto. 2024.
\newblock Length-controlled alpacaeval: A simple way to debias automatic evaluators.
\newblock \emph{arXiv preprint arXiv:2404.04475}.

\bibitem[{Fabbri et~al.(2021)Fabbri, Kry{\'s}ci{\'n}ski, McCann, Xiong, Socher, and Radev}]{fabbri2021summeval}
Alexander~R Fabbri, Wojciech Kry{\'s}ci{\'n}ski, Bryan McCann, Caiming Xiong, Richard Socher, and Dragomir Radev. 2021.
\newblock Summeval: Re-evaluating summarization evaluation.
\newblock \emph{Transactions of the Association for Computational Linguistics}, 9:391--409.

\bibitem[{Frantar et~al.(2022)Frantar, Ashkboos, Hoefler, and Alistarh}]{frantar2022gptq}
Elias Frantar, Saleh Ashkboos, Torsten Hoefler, and Dan Alistarh. 2022.
\newblock Gptq: Accurate post-training quantization for generative pre-trained transformers.
\newblock \emph{arXiv preprint arXiv:2210.17323}.

\bibitem[{Fu et~al.(2022)Fu, Chen, Laskar, Gardiner, Hiranandani, and Tn}]{fu2022entity}
Xue-Yong Fu, Cheng Chen, Md~Tahmid~Rahman Laskar, Shayna Gardiner, Pooja Hiranandani, and Shashi~Bhushan Tn. 2022.
\newblock Entity-level sentiment analysis in contact center telephone conversations.
\newblock \emph{arXiv preprint arXiv:2210.13401}.

\bibitem[{Guo and Vosoughi(2023)}]{guo2023length}
Xiaobo Guo and Soroush Vosoughi. 2023.
\newblock Length does matter: Summary length can bias summarization metrics.
\newblock In \emph{Proceedings of the 2023 Conference on Empirical Methods in Natural Language Processing}, pages 15869--15879.

\bibitem[{Gururangan et~al.(2018)Gururangan, Swayamdipta, Levy, Schwartz, Bowman, and Smith}]{gururangan2018annotation}
Suchin Gururangan, Swabha Swayamdipta, Omer Levy, Roy Schwartz, Samuel~R Bowman, and Noah~A Smith. 2018.
\newblock Annotation artifacts in natural language inference data.
\newblock \emph{arXiv preprint arXiv:1803.02324}.

\bibitem[{Hendry et~al.(2021)Hendry, Darari, Nurfadillah, Khanna, Sun, Condylis, and Taufik}]{hendry2021topic}
Darell Hendry, Fariz Darari, Raditya Nurfadillah, Gaurav Khanna, Meng Sun, Paul~Constantine Condylis, and Natanael Taufik. 2021.
\newblock Topic modeling for customer service chats.
\newblock In \emph{2021 International Conference on Advanced Computer Science and Information Systems (ICACSIS)}, pages 1--6. IEEE.

\bibitem[{Hu et~al.(2021)Hu, Shen, Wallis, Allen-Zhu, Li, Wang, Wang, and Chen}]{hu2021lora}
Edward~J Hu, Yelong Shen, Phillip Wallis, Zeyuan Allen-Zhu, Yuanzhi Li, Shean Wang, Lu~Wang, and Weizhu Chen. 2021.
\newblock Lora: Low-rank adaptation of large language models.
\newblock \emph{arXiv preprint arXiv:2106.09685}.

\bibitem[{Khasanova et~al.(2022)Khasanova, Hiranandani, Gardiner, Chen, Fu, and Corston-Oliver}]{khasanova2022developing}
Elena Khasanova, Pooja Hiranandani, Shayna Gardiner, Cheng Chen, Xue-Yong Fu, and Simon Corston-Oliver. 2022.
\newblock Developing a production system for purpose of call detection in business phone conversations.
\newblock \emph{arXiv preprint arXiv:2205.06904}.

\bibitem[{Khobragade et~al.(2019)Khobragade, Patel, Namdev, Mishra, and Bhattacharyya}]{khobragade2019machine}
Rakesh Khobragade, Heaven Patel, Anand Namdev, Anish Mishra, and Pushpak Bhattacharyya. 2019.
\newblock Machine translation evaluation using bi-directional entailment.
\newblock \emph{arXiv preprint arXiv:1911.00681}.

\bibitem[{Kwon et~al.(2023)Kwon, Li, Zhuang, Sheng, Zheng, Yu, Gonzalez, Zhang, and Stoica}]{kwon2023efficientmemorymanagementlarge}
Woosuk Kwon, Zhuohan Li, Siyuan Zhuang, Ying Sheng, Lianmin Zheng, Cody~Hao Yu, Joseph~E. Gonzalez, Hao Zhang, and Ion Stoica. 2023.
\newblock \href {https://arxiv.org/abs/2309.06180} {Efficient memory management for large language model serving with pagedattention}.
\newblock \emph{Preprint}, arXiv:2309.06180.

\bibitem[{Liu et~al.(2022)Liu, Fabbri, Liu, Zhao, Nan, Han, Han, Joty, Wu, Xiong et~al.}]{liu2022revisiting}
Yixin Liu, Alexander~R Fabbri, Pengfei Liu, Yilun Zhao, Linyong Nan, Ruilin Han, Simeng Han, Shafiq Joty, Chien-Sheng Wu, Caiming Xiong, et~al. 2022.
\newblock Revisiting the gold standard: Grounding summarization evaluation with robust human evaluation.
\newblock \emph{arXiv preprint arXiv:2212.07981}.

\bibitem[{McInnes et~al.(2017)McInnes, Healy, Astels et~al.}]{mcinnes2017hdbscan}
Leland McInnes, John Healy, Steve Astels, et~al. 2017.
\newblock hdbscan: Hierarchical density based clustering.
\newblock \emph{J. Open Source Softw.}, 2(11):205.

\bibitem[{Moulavi et~al.(2014)Moulavi, Jaskowiak, Campello, Zimek, and Sander}]{moulavi2014density}
Davoud Moulavi, Pablo~A Jaskowiak, Ricardo~JGB Campello, Arthur Zimek, and J{\"o}rg Sander. 2014.
\newblock Density-based clustering validation.
\newblock In \emph{Proceedings of the 2014 SIAM international conference on data mining}, pages 839--847. SIAM.

\bibitem[{Pan et~al.(2024)Pan, Wu, Jiang, Xia, Luo, Zhang, Lin, Ruhle, Yang, Lin, Zhao, Qiu, and Zhang}]{pan-etal-2024-llmlingua}
Zhuoshi Pan, Qianhui Wu, Huiqiang Jiang, Menglin Xia, Xufang Luo, Jue Zhang, Qingwei Lin, Victor Ruhle, Yuqing Yang, Chin-Yew Lin, H.~Vicky Zhao, Lili Qiu, and Dongmei Zhang. 2024.
\newblock \href {https://aclanthology.org/2024.findings-acl.57} {{LLML}ingua-2: Data distillation for efficient and faithful task-agnostic prompt compression}.
\newblock In \emph{Findings of the Association for Computational Linguistics ACL 2024}, pages 963--981, Bangkok, Thailand and virtual meeting. Association for Computational Linguistics.

\bibitem[{Papadia et~al.(2022)Papadia, Pacella, and Giliberti}]{papadia2022topic}
Gabriele Papadia, Massimo Pacella, and Vincenzo Giliberti. 2022.
\newblock Topic modeling for automatic analysis of natural language: A case study in an italian customer support center.
\newblock \emph{Algorithms}, 15(6):204.

\bibitem[{Papineni et~al.(2002)Papineni, Roukos, Ward, and Zhu}]{papineni2002bleu}
Kishore Papineni, Salim Roukos, Todd Ward, and Wei-Jing Zhu. 2002.
\newblock Bleu: a method for automatic evaluation of machine translation.
\newblock In \emph{Proceedings of the 40th annual meeting of the Association for Computational Linguistics}, pages 311--318.

\bibitem[{Reimers and Gurevych(2019)}]{reimers-2019-sentence-bert}
Nils Reimers and Iryna Gurevych. 2019.
\newblock \href {http://arxiv.org/abs/1908.10084} {Sentence-bert: Sentence embeddings using siamese bert-networks}.
\newblock In \emph{Proceedings of the 2019 Conference on Empirical Methods in Natural Language Processing}. Association for Computational Linguistics.

\bibitem[{Tahmid Rahman~Laskar et~al.(2022)Tahmid Rahman~Laskar, Chen, Fu, and Bhushan~TN}]{tahmid2022improving}
Md~Tahmid Rahman~Laskar, Cheng Chen, Xue-Yong Fu, and Shashi Bhushan~TN. 2022.
\newblock Improving named entity recognition in telephone conversations via effective active learning with human in the loop.
\newblock \emph{arXiv e-prints}, pages arXiv--2211.

\bibitem[{Tang et~al.(2003)Tang, Pellom, and Hacioglu}]{tang2003call}
Min Tang, Bryan Pellom, and Kadri Hacioglu. 2003.
\newblock Call-type classification and unsupervised training for the call center domain.
\newblock In \emph{2003 IEEE Workshop on Automatic Speech Recognition and Understanding (IEEE Cat. No. 03EX721)}, pages 204--208. IEEE.

\bibitem[{Vaswani et~al.(2017)Vaswani, Shazeer, Parmar, Uszkoreit, Jones, Gomez, Kaiser, and Polosukhin}]{vaswani2017attention}
Ashish Vaswani, Noam Shazeer, Niki Parmar, Jakob Uszkoreit, Llion Jones, Aidan~N Gomez, {\L}ukasz Kaiser, and Illia Polosukhin. 2017.
\newblock Attention is all you need.
\newblock \emph{Advances in neural information processing systems}, 30.

\bibitem[{Zou et~al.(2021)Zou, Zhao, Kang, Lin, Peng, Jiang, Sun, Zhang, Huang, and Liu}]{zou2021topic}
Yicheng Zou, Lujun Zhao, Yangyang Kang, Jun Lin, Minlong Peng, Zhuoren Jiang, Changlong Sun, Qi~Zhang, Xuanjing Huang, and Xiaozhong Liu. 2021.
\newblock Topic-oriented spoken dialogue summarization for customer service with saliency-aware topic modeling.
\newblock In \emph{Proceedings of the AAAI Conference on Artificial Intelligence}, 16, pages 14665--14673.

\end{thebibliography}

\appendix



\end{document}